# An image processing of a Raphael's portrait of Leonardo

**Amelia Carolina Sparavigna**
Dipartimento di Fisica
Politecnico di Torino

In one of his paintings, the School of Athens, Raphael is depicting Leonardo da Vinci as the philosopher Plato. Some image processing tools can help us in comparing this portrait with two Leonardo's portraits, considered as self-portraits.

The "Scuola di Atene" is one of the most famous paintings by Raphael, the Italian Renaissance artist. Painted between 1510 and 1511, this fresco decorates the wall of one of the rooms, the "Stanza della Segnatura", in the Apostolic Palaces of Vatican. The great Greek philosophers are represented inside a classic architecture. At the central position of this masterpiece, we see two philosophers, Plato on the left and Aristotle, his student, on the right. Plato is shown as a wise-looking man (see Fig.1). It is believed that Raphael based the Plato's face on the features of Leonardo da Vinci [1]. The two artists probably had established a direct interaction when Raphael spent a period of his life in Florence, perhaps from about 1504 to 1508 [2-4]. Leonardo da Vinci returned to Florence from 1500 to 1506: therefore, if the image of Plato is a portrait of Leonardo, this means that Raphael depicted him when Leonardo was 52 or 54 year old.

There is a portrait in red chalk, dated approximately 1510 and held at the Biblioteca Reale of Turin, which is widely accepted as a self-portrait of Leonardo da Vinci. It is thought that Leonardo drew this self-portrait at the age of 58 or 60 (see Fig.2). Ref.5 tells that this well-known drawing is not universally accepted as a self-portrait, because the depicted face appears to be quite old, suggesting that Leonardo represented his father or grandfather. Another possibility is that Leonardo altered himself, in order that Raphael might use it for his Plato. However, Plato does not look so old in the painting by Raphael.

In any case, let us try to find some matching points between the portraits, that of the man in red chalk (Fig.2) - let us call it, from now on, the self-portrait in red chalk – and the image of Plato (Fig.1) that Raphael had depicted in his fresco. To match the two faces, the image processing is fundamental: in particular, we will use another Leonardo's portrait merged with the self-portrait in red chalk. This portrait is a drawing of the Codex on the Flight of Birds, which Leonardo had partially hidden by his writing, as shown in Fig3, left panel. According to Carlo Pedretti, an Italian historian expert on the life and works of Leonardo, this is a self-portrait [6,7] made when the artist was young. The codex dated approximately 1505, but the portrait is older for sure: Leonardo recycled the paper for the composition of the Codex.

To use this portrait it is necessary to remove the written text. Carlo Pedretti was the first to suggest a "restoration" of this drawing [7], of course not of the real page of the Codex, but made on a photographic plate. The result that Pedretti obtained in 1975, with a negative-positive photographic procedure, was quite good. However, it was just in 2009 that the portrait became popular because of an Italian scientific journalist, Piero Angela, that presented a digital restoration of this portrait [8,9], that is, a restoration of the corresponding digital image. In 2009, I have proposed a simple approach that uses an iterative procedure based on thresholding and interpolation with nearest neighbouring pixels [10,11]. Recently, I proposed a further processing with a wavelet-filtering program, Iris [12-14]: the result is shown in Fig.3, right panel. According to Pedretti, this is the young Leonardo da Vinci self-portrait.

For any comparison with the Raphael's portrait, we have to complete this image, since the artist abandoned it unfinished. We use another processing tool, the GIMP [15], for this purpose. Using GIMP, we can add this drawing of the young man to the self-portrait in red chalk of the old man. The result is given in Fig.4: besides showing that the two faces have the same relative distances of eyes, nose and mouth, this portrait makes the old Leonardo look younger.

In Figure 5 we have the two images, the Raphael painting on the right and the result of merging the two Leonardo's drawings on the left, shown side by side. Let me remark that we are looking at two images obtained from originals created by artists who used different techniques and a different rendering of the head position. Moreover, there is another fact, which is in my opinion quite important, that the two portraits are showing a distinct side of the face. And we know very well that the two sides are not equal and that the existing small differences create the "good" and "bad" side of our faces [16].

Let us remember that for all the living creatures, the bilateral symmetry [17] of the body is an approximate symmetry: the two halves, left and right, of the body and then of the face, are not perfectly symmetrical. The symmetry of human faces is a subject of several studies. Some researchers are supporting the idea that more symmetry means more beauty and freedom from diseases [18-20]. On the other hand, a face, which is too symmetric, gives the impression of being unnatural [21].

The fact that the two sides are different is quite relevant if we are comparing a self-portrait with a portrait, because we must be sure to compare the same side of the face. For the explanation, let me use Fig.6. Let us consider two canvasses, having on them a self-portrait and a portrait, with the head depicted in the same position, the two paintings are showing a different side of the face. When the artist is depicting a self-portrait, he is looking at the face in a mirror. When it is another artist depicting the portrait, he is looking at the face directly. For this reason, if the face on the canvas has the same position, the depicted sides turn out to be different. Therefore, if the left image of Fig.5 is a self-portrait and the right image is a portrait, it is necessary to reflect one of then, to point out that we are seeing different sides.

I decided to change the Raphael's image, with a reflection and a small rotation using GIMP. Moreover, I converted the colours in grey tones, to avoid the vision of different hues. Fig.7 gives the result. Is the figure showing the same person? I guess that there is this possibility, but further studies are necessary. Let me then avoid a direct answer and just write some conclusions.

Using the image processing we had compared portraits having quite different origins. This is telling that several processing tools, some of them freely available, can help in the study of history and arts. For what concerns the specific case, it seems from Fig.7, that the structure of the two faces, in particular of nose and cheekbones, is quite similar. We can also see that one of the eyes is a little bit larger in both images. According on the previous discussion on portrait and self-portrait (Fig.6), I tend to consider the Raphael's Plato based on a direct interaction between Raphael and Leonardo, when Raphael was in Florence, and then on a previous portrait or drawing that Raphael made of Leonardo.

**References**
1. Raffaello, presentato da Mchele Prisco, Milano, Rizzoli Editore, 1966; Raffaello Sanzio, presentato da M.G. Ciardi Dupré, Milano, Fratelli Fabbri Editore, 1963.
2. Cecil Gould, The Sixteenth Century Italian Schools, National Gallery Catalogues, London 1975; Henry Strachey, Raphael, G. Bell and Sons, London, 1911.
3. Raphael, http://en.wikipedia.org/wiki/Raphael,
4. The School of Athens, http://en.wikipedia.org/wiki/The_School_of_Athens
5. Cultural depictions of Leonardo da Vinci, http://en.wikipedia.org/wiki/Cultural_depictions_of_Leonardo_da_Vinci
6.. E. Crispino, C. Pedretti, C. Frost, Leonardo: Art and Science, Giunti, 2001.
7. C. Pedretti, Disegni di Leonardo da Vinci e della sua scuola alla Biblioteca Reale di Torino, Giunti Barbera, Firenze, 1975; C. Pedretti, A Chronology of Leonardo Da Vinci's Architectural Studies after 1500, E. Droz, Geneva, 1962.
8. ANSA.it - News in English - Leonardo self- portrait 'discovered', 2009 and also BBC NEWS Europe - 'Early Leonardo portrait' found, 2009.
9. http://www.leonardo3.net/
10. Amelia Carolina Sparavigna, 2009, The Digital Restoration of Da Vinci's Sketches, http://arxiv.org/abs/0903.1448
11. Amelia Carolina Sparavigna, 2009, Digital Restoration of Ancient Papyri, http://arxiv.org/abs/0903.5045
12. Amelia Carolina Sparavigna, 2011, A self-portrait of young Leonardo, http://arxiv.org/abs/1111.4654
13. Iris © 1999-2010, Christian Buil, http://www.astrosurf.com/buil/us/iris/iris.htm
14. Amelia Carolina Sparavigna, 2009, Enhancing the Google imagery using a wavelet filter, http://arxiv.org/abs/1009.1590
15. GIMP © 2001-2011, http://www.gimp.org/
16. I have read on the Glamour Magazine about a simple experiment by P. Gugliemetti, Do You Have A Good Side And Bad Side Of Your Face?, 11-13-2008. The author writes "At a party over the summer, I mentioned to someone how I have a good side and bad side, and she thought I was just being dramatic. So I had her take a photo of each side and we showed the shots to random people in the room, asking them to vote on which side was my prettier one. Every single person voted right! Then we tried this on other people, lining them up one-by-one against a white wall, shooting their sides, and having people vote. Only a couple had equally attractive sides."
17. Bilateral symmetry of a body means that there exists a plane which is dividing the body into two mirror image halves. An operation of reflection shows that the two halves coincide.
18. G. Rhodes and L.A. Zebrowitz, Facial Attractiveness - Evolutionary, Cognitive, and Social Perspectives. Ablex. ISBN 1567506364, 2002
19. R.J. Edler, Journal of Orthodontics Vol.28(2), pag.159, 2001
20. K. Grammer and R. Thornhill, Journal of Comparative Psychology, Vol. 108, pag.233, 1994.
21. R. Kowner, Journal of Experimental Psychology: Human Perception and Performance, Vol.22, pag.662, 1996.

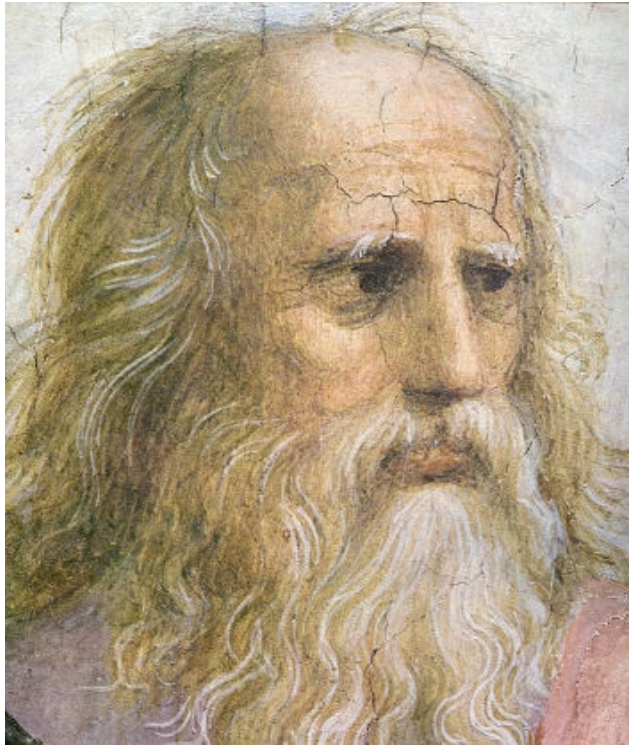

Fig.1 Raphael's Plato (adapted from an image at http://www.aiwaz.net/)

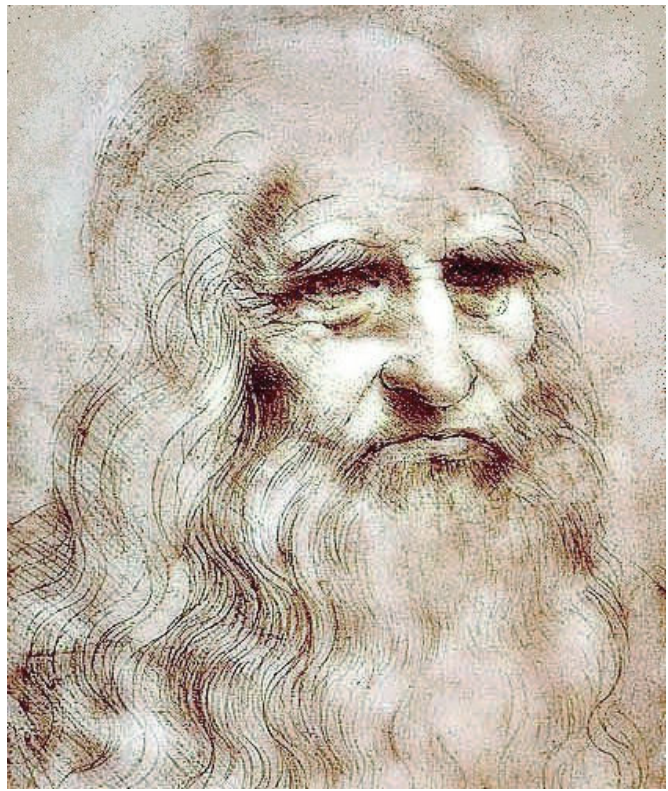

Fig.2 Leonardo's portrait in red chalk (dated approx. 1510) held at the Biblioteca Reale of Turin.

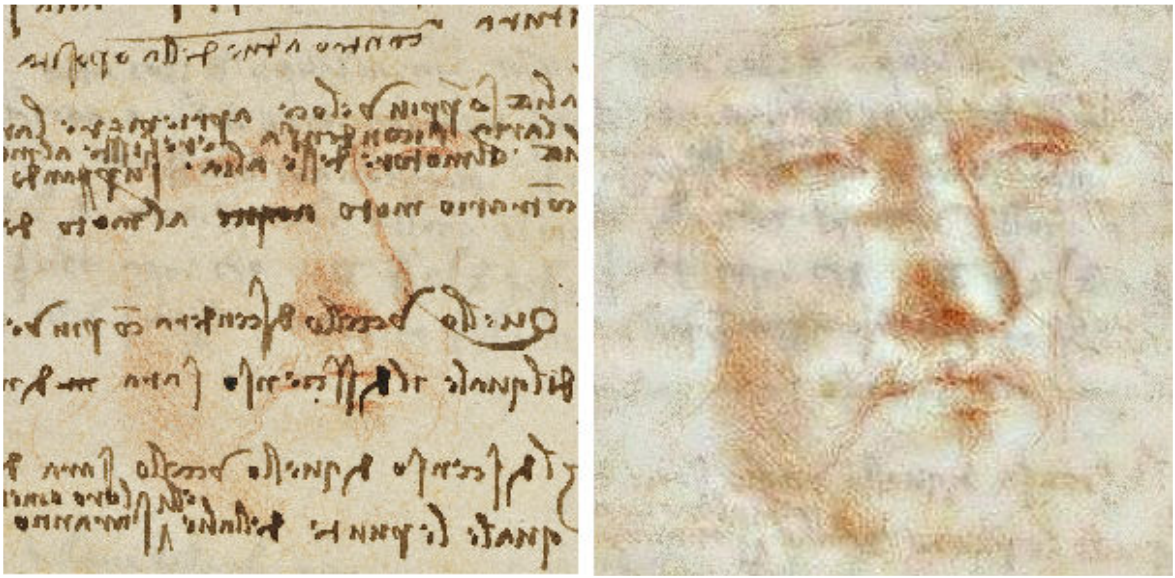

Fig.3. A page of the Codex on the Flight of Birds contains a Leonardo's portrait. Using a digital restoration that removes the writing, the portrait appears.

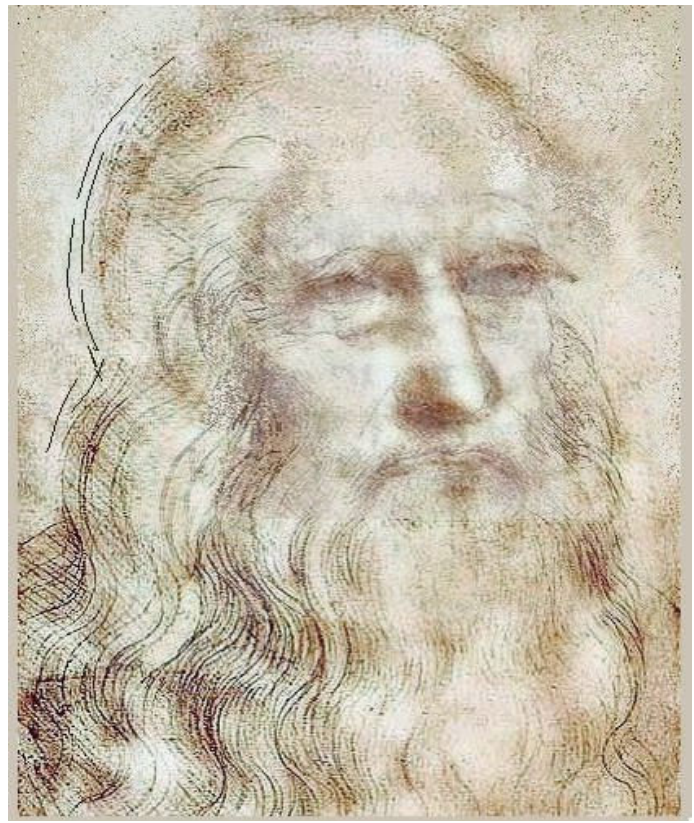

Fig.4 Using GIMP [15] we can add the portrait of the young man (Fig.3, right) to the self-portrait in red chalk (Fig.2) of the old man.

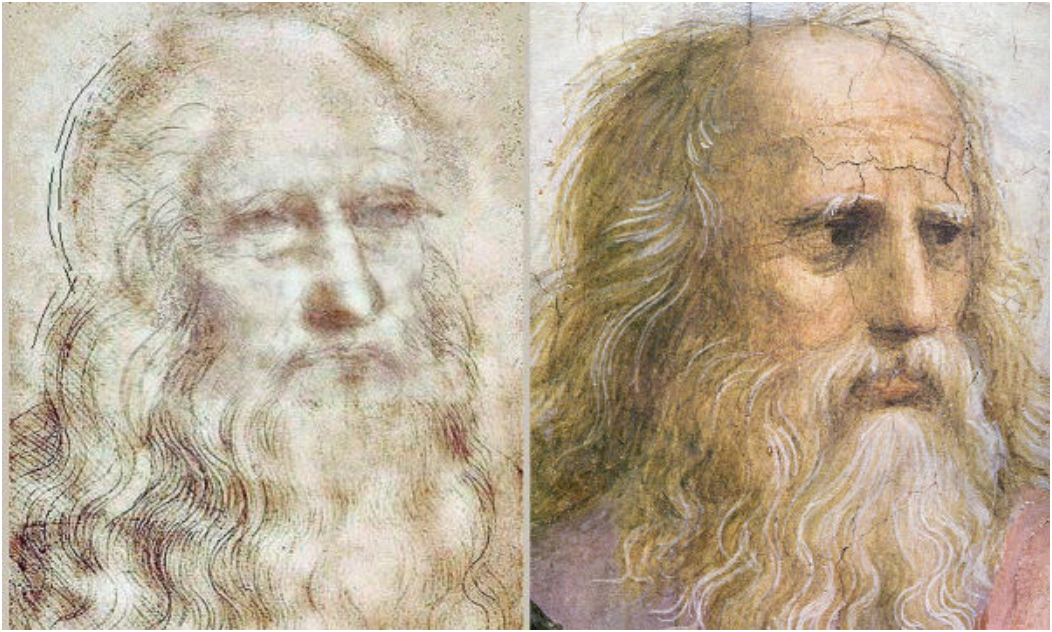

Fig.5 On the right, the Raphael painting and on the left, the result of a merging of two Leonardo's drawings

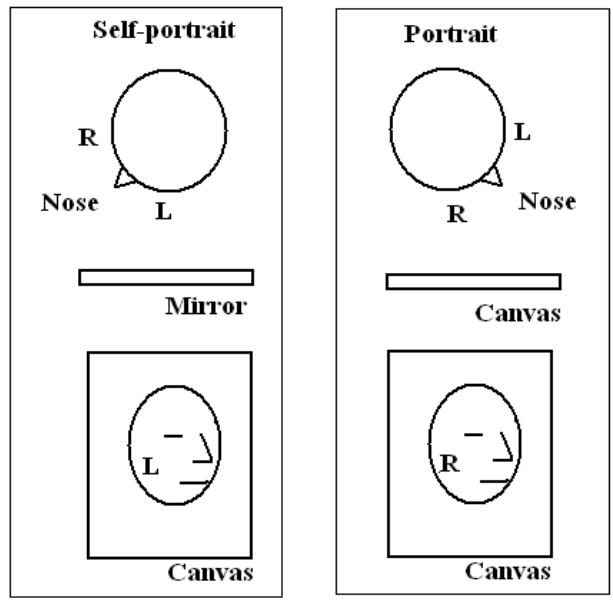

Fig.6. Let us consider two canvasses, having on them a self-portrait and a portrait respectively, with the head depicted in the same position. The side of the face is different. When an artist is depicting a self-portrait, he is looking at the face in a mirror. Assuming the position of the head as above, the self-portrait is showing the left side of the face. In the case that it is another artist depicting the portrait, he is looking at the face directly, and then the side depicted is the right one.

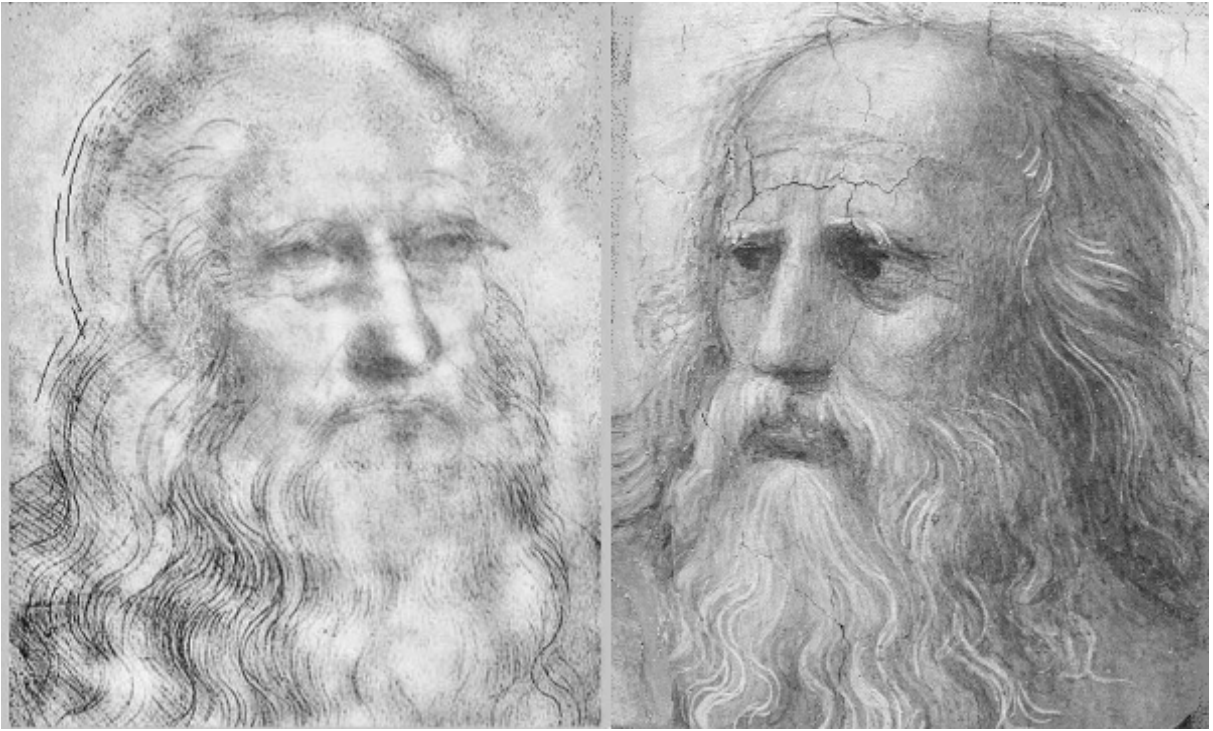

Fig.7 Is this the same person?